\documentclass[runningheads]{llncs}
\usepackage[T1]{fontenc}
\usepackage{graphicx}
\usepackage{amsmath,amssymb,amsfonts}
\usepackage{hyperref}
\usepackage{pifont}
\usepackage[misc]{ifsym}
\hypersetup{
    colorlinks=true,  
    linkcolor=blue,       
    citecolor=blue,    
    filecolor=blue, 
    urlcolor=blue 
}

\begin{document}

\title{Adjustable Robust Transformer for High Myopia Screening in Optical Coherence Tomography}

\titlerunning{Adjustable Robust Transformer for High Myopia Screening}

\author{Xiao Ma\inst{1} \and
Zetian Zhang\inst{1} \and
Zexuan Ji\inst{1} \and
Kun Huang\inst{1} \and
Na Su\inst{2} \and
Songtao	Yuan\inst{2} \and
Qiang Chen\inst{1}$^{(\textrm{\Letter})}$}

\authorrunning{Xiao Ma et al.}

\institute{School of Computer Science and Engineering, Nanjing University of Science and Technology, Nanjing, China\\
\email{chen2qiang@njust.edu.cn} \and
Department of Ophthalmology, The First Affiliated Hospital of Nanjing Medical University, Nanjing, China}

\maketitle              
% typeset the header of the contribution

\begin{abstract}
Myopia is a manifestation of visual impairment caused by an excessively elongated eyeball. Image data is critical material for studying high myopia and pathological myopia. Measurements of spherical equivalent and axial length are the gold standards for identifying high myopia, but the available image data for matching them is scarce. In addition, the criteria for defining high myopia vary from study to study, and therefore the inclusion of samples in automated screening efforts requires an appropriate assessment of interpretability. In this work, we propose a model called adjustable robust transformer (ARTran) for high myopia screening of optical coherence tomography (OCT) data. Based on vision transformer, we propose anisotropic patch embedding (APE) to capture more discriminative features of high myopia. To make the model effective under variable screening conditions, we propose an adjustable class embedding (ACE) to replace the fixed class token, which changes the output to adapt to different conditions. Considering the confusion of the data at high myopia and low myopia threshold, we introduce the label noise learning strategy and propose a shifted subspace transition matrix (SST) to enhance the robustness of the model. Besides, combining the two structures proposed above, the model can provide evidence for uncertainty evaluation. The experimental results demonstrate the effectiveness and reliability of the proposed method. Code is available at: \url{https://github.com/maxiao0234/ARTran}.

\end{abstract}

\section{Introduction}

\begin{figure}
\centering
\includegraphics[width=0.85\textwidth]{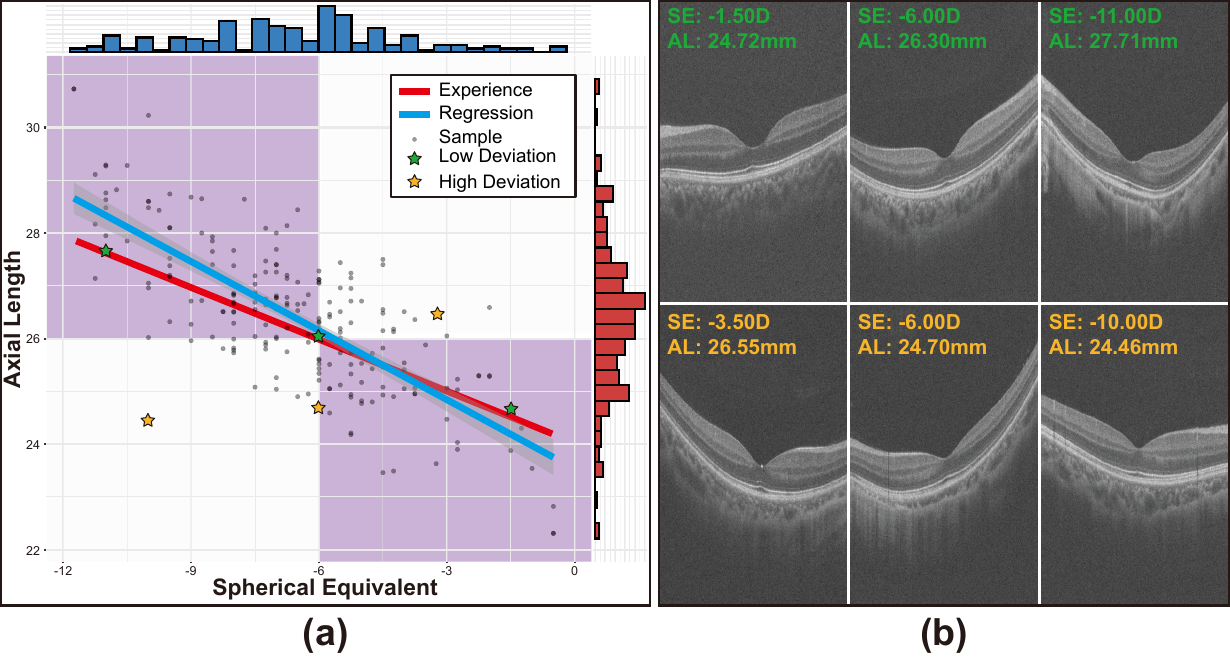}
\caption{Illustration of spherical equivalent (SE) and axial length (AL) for some samples in the dataset. (a) Scatterplot on two measured values. The red line indicates the rough estimation relationship from clinical experience, the blue line indicates the linear representation of regression. (b) OCT examples for low deviation of the experience trend (the top row) and with high deviation (the bottom row).} 
\label{fig-1}
\end{figure}

Myopia, resulting in blurred distance vision, is one of the most common eye diseases, with a rising prevalence around the world, particularly among schoolchildren and young adults \cite{morgan2012myopia,holden2016global,resnikoff2019myopia}. Common wisdom attributes the causes of myopia to excessive elongation of the eyeball, the development of which can continue throughout childhood, and especially in patients with high myopia, throughout adulthood \cite{baird2020myopia,bullimore2021risks}. In the coming decades, the prognosis for patients with high myopia will continue to deteriorate, with some developing pathological myopia, leading to irreversible vision damage involving posterior scleral staphyloma, macular retinoschisis, macular hole, retinal detachment, choroidal neovascularization, dome-shaped macula, etc. \cite{buch2004prevalence,iwase2006prevalence,choudhury2018prevalence,ruiz2019myopic}.

As a crucial tool in the study of high myopia, fundus images demonstrate the retinal changes affected by myopia. Myopic maculopathy in color fundus photographs (CFP) can be important evidence in the evaluation of high myopia and myopic fundus diseases \cite{ohno2015international}. However, some myopic macular lesions such as myopic traction maculopathy and domeshaped macula are usually not observed in CFP. Optical coherence tomography (OCT), characterized by non-invasive and high-resolution three-dimensional retinal structures, has more advantages in the examination of high myopia \cite{fang2019oct,li2022advances}. Several studies have used convolutional neural networks to automatically diagnose high myopia and pathological myopia \cite{choi2021deep,li2022development}. Choi {\em et al.} employed two ResNet-50 \cite{he2016deep} networks to inference vertical and horizontal OCT images simultaneously. Li {\em et al.} introduced focal loss into Inception-Resnet-V2 \cite{szegedy2017inception} to enhance its identification ability. However, the different classes of images in these tasks have significant differences, and the performance of the model, when used for more complex screening scenarios, is not discussed. Hence, this work aims to design a highly generalizable screening model for high myopia on OCT images.

There exist challenges to developing an automatic model that meets certain clinical conditions. For the inclusion criteria for high myopia, different studies will expect different outputs under different thresholds. High myopia is defined by a spherical equivalent (SE) $\le -6.0$ dioptres ($D$) or an axial length (AL) $\ge 26.0mm$ in most cases, but some researchers set the threshold of SE to $-5.0 D$ \cite{bullimore2019myopia} or $-8.0 D$ \cite{nakao2022quantitative}. Moreover, some scenarios where screening risk needs to be controlled may modify the thresholds appropriately. To remedy this issue, the screening scheme should ideally output compliant results for different thresholds. For the accuracy of supervision, although a worse SE has a higher chance of causing structural changes in the retina (and vice versa), the two are not the single cause and effect, i.e., there are natural label noises when constructing datasets. One direct piece of evidence is that both measures of SE and AL can be used as inclusion criteria for high myopia, but there is disagreement in some samples. We illustrate the distribution of samples with both SE and AL in our dataset in Fig.~\ref{fig-1}(a). Clinical experience considers that AL and SE can be roughly estimated from each other using a linear relationship, which is indicated by the red line in Fig.~\ref{fig-1}(a). Most of the samples are located in the area (purple) that satisfies both conditions, but the remaining samples can only satisfy one. In more detail, Fig.~\ref{fig-1}(b) shows three samples with a low deviation of the experience trend (the top row) and three samples with a high deviation (the bottom row), where the worse SE does not always correspond to longer AL or more retinal structural changes. To mitigate degenerate representation caused by noisy data, the screening scheme should avoid over-fitting of extremely confusing samples. Besides, rich interpretable decisions support enhancing confidence in screening results. To this end, the screening scheme should evaluate the uncertainty of the results.

The contributions of our work are summarized as: (1) we propose a novel adjustable robust transformer (ARTran) for high myopia screening to adapt variable inclusion criteria. We design an anisotropic patch embedding (APE) to encode more myopia-related information in OCT images, and an adjustable class embedding (ACE) to obtain adjustable inferences. (2) We propose shifted subspace transition matrix (SST) to mitigate the negative impacts of label noise, which maps the class-posteriors to the corresponding distribution range according to the variable inclusion criteria. (3) We implement our ARTran on a high myopia dataset and verify the effectiveness of screening, and jointly use the proposed modules to generate multi-perspective uncertainty evaluation results.

\section{Method}

In this section, we propose a novel framework for high myopia screening in OCT called adjustable robust transformer (ARTran). This model can obtain the corresponding decisions based on the given threshold of inclusion criteria for high myopia and is trained end-to-end for all thresholds at once. During the testing phase, the screening results can be predicted interactively for a given condition.

\begin{figure}
\includegraphics[width=\textwidth]{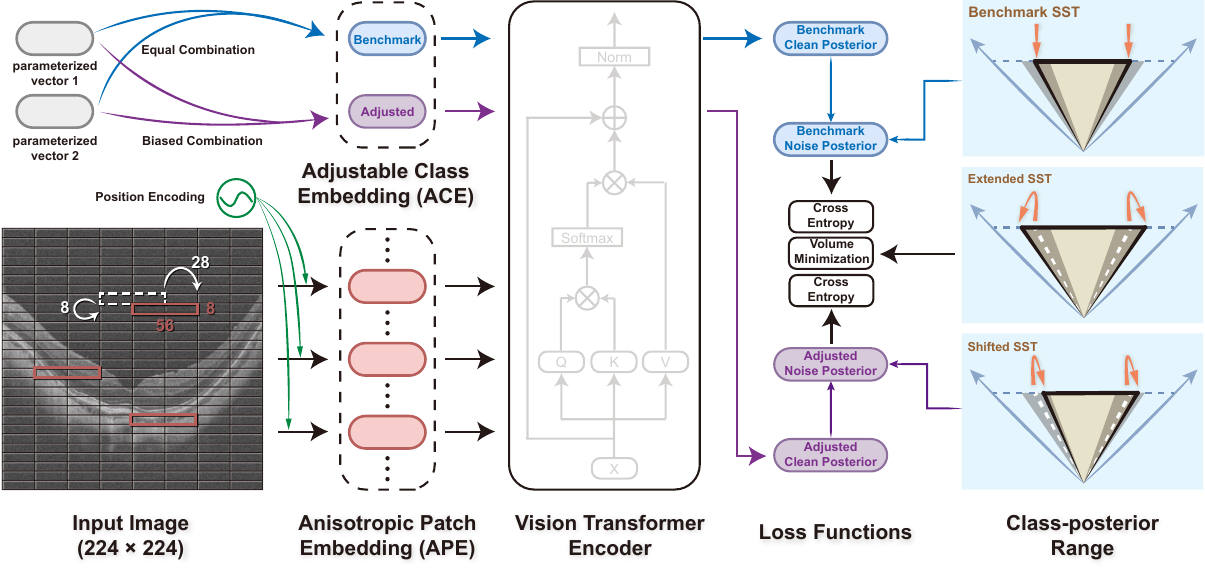}
\caption{The framework of our proposed adjustable robust transformer (ARTran). The proposed APE and ACE encode the image information and adjustment information as the input to transformer. The proposed SST learns the label noise and establishes the connection between the class-posteriors. The range of class-posteriors are also shown.} 
\label{fig-2}
\end{figure}

\subsection{Modified Vision Transformer for Screening}

Transformers have shown promising performance in visual tasks attributed to long-range dependencies. Inspired by this, we use ViT \cite{dosovitskiy2020image} as the backbone of our framework and make improvements for the task of screening high myopia OCT images. 

Patients with high myopia are usually accompanied by directional structural changes, such as increased retinal curvature and choroidal thinning. On the other hand, due to the direction of light incidence perpendicular to the macula, OCT images have unbalanced information in the horizontal and vertical directions. Therefore, we propose a novel non-square strategy called anisotropic patch embedding (APE) to replace vanilla patch embedding for perceiving finer structural information, where an example is shown in Fig.~\ref{fig-2}. First, we reduced the height size of the patches. This strategy increases the number of patches per column, i.e., the number of patches within the retinal height range, which enhances the information perception of overall and individual layer thickness. In order not to introduce extra operations, we reduce the number of patches per row. We also use an overlapping sampling strategy for sliding windows in the horizontal direction, where a wider patch captures more information about the peripheral structure of the layer in which it is located. Specifically, based on the $ 16 \times 16 $ size of ViT, our APE changed the height to 8 pixels and the width to 56 pixels with an overlap of 28 pixels, which keeps the number of embedded patches.

Considering that different researchers may have different requirements for inclusion criteria or risk control, we propose adjustable class embedding (ACE) to adapt to variable conditions. Take the $-6.0D$ as the benchmark of inclusion criteria, we define the relationship of the biased label, SE, and the adjustment coefficient $\Delta$:
\begin{equation}
    Y(SE=s,\Delta=\delta) :=
    \begin{cases}
        0 & ,-0.25D \cdot \delta + s > -6.0D \\
        1 & ,-0.25D \cdot \delta + s \le -6.0D \\
    \end{cases}
    , -1 \le \delta \le 1
\end{equation}
where $1$ indicates the positive label and $0$ indicates the negative label. Our ACE introduces two parameterized vectors $v_1$ and $v_2$, and constructs a linear combination with the given $\delta$ to obtain $v_{APE}(\delta)$ to replace the fixed class token:
\begin{equation}
    v_{APE}(\delta) = \frac{1 + \delta}{2} \cdot v_1 + \frac{1 - \delta}{2} \cdot v_2
\end{equation}
where $v_{APE}(0)$ is the equal combination and others are biased combinations. Several studies have demonstrated impressive performance in multi-label tasks using transformers with multiple class tokens \cite{lanchantin2021general,xu2022multi}. Inspired by this, we set $v_{APE}(0)$ as the benchmark class embedding and $v_{APE}(\delta)$ as the biased class embedding. In the training stage, the inclusion threshold is varied around the benchmark, affecting the supervision consequently. The ACE adaptively changes the input state according to the scale of the adjustment coefficient to obtain the corresponding output. The scheme for constructing loss functions using two outputs is described in Section \ref{SST}. In the testing stage, the ACE interactively makes the model output different results depending on the given conditions. Furthermore, we did not add the position encoding to ACE, so both tokens are position-independently involved to multi-head self-attention and are only distinguished by adjustment coefficient.

\subsection{Shifted Subspace Transition Matrix}\label{SST}

To enhance the robustness of the model, we follow the common assumptions of some impressive label noise learning methods\cite{xia2019anchor,yao2020dual}. Conventional wisdom is that the clean class-posterior $P(Y|X=x)$ can be inferred by utilizing the noisy class-posterior $P(\tilde{Y}|X=x)$ and the transition matrix $\textbf{\emph{T}}(x)$:
\begin{equation}
    P(\tilde{\textbf{\emph{Y}}}|X=x) = \textbf{\emph{T}}(x)P(\textbf{\emph{Y}}|X=x)
\end{equation}
where the transition matrix guarantees statistical consistency. Li {\em et al.} \cite{li2021provably} have proved that identifying the transition matrix can be treated as the problem of recovering simplex for any $x$, i.e., $\textbf{\emph{T}}(x)=\textbf{\emph{T}}$. Based on this theory, in this work, we further propose a class-dependent and adjustment-dependent transition matrix $\textbf{\emph{T}}(x,\delta)=\textbf{\emph{T}}(\delta)$ called shifted subspace transition matrix (SST) to adapt to the variable class-posteriors distribution space. 

Simplistically, this work only discusses the binary classification approach applicable to screening. For each specific inclusion threshold $\delta \in [-1, 1]$, the range of the noisy class-posterior is determined jointly by the threshold and the SST:
\begin{equation}
\begin{bmatrix}
P(\tilde{Y}=0|x,\delta)  \\
P(\tilde{Y}=1|x,\delta)
\end{bmatrix}
=
\begin{bmatrix}
\textbf{\emph{T}}_{1,1}(\delta)   &  1-\textbf{\emph{T}}_{2,2}(\delta) \\
1-\textbf{\emph{T}}_{1,1}(\delta) &  \textbf{\emph{T}}_{2,2}(\delta)
\end{bmatrix}
\begin{bmatrix}
P(Y=0|x,\delta)  \\
P(Y=1|x,\delta)
\end{bmatrix}
\end{equation}
where $T_{i,i}(\delta) > 0.5$ is the $i_{th}$ diagonal element of SST, and the sum of each column of the matrix is $1$. Thus any class-posterior of $x$ is inside the simplex form columns of $\textbf{\emph{T}}(\delta)$ \cite{boyd2004convex}. As shown in Fig~\ref{fig-2}, the benchmark simplex is formed by $\textbf{\emph{T}}(0)$, where the orange arrows indicate the two sides of the simplex. When adjusting the inclusion criteria for high myopia, we expect the adjusted class-posterior to prefer same the adjustment direction compared to the benchmark. One solution is to offset both the upper and lower bounds of the class-posterior according to the scales of the adjustment coefficient:
\begin{equation}
\textbf{\emph{T}}_{i,i}(\delta) = \frac{1 + S(\theta_{0}) \cdot S(\theta_{i})}{2} + \frac{1 - S(\theta_{0})}{4}  \cdot
    \begin{cases}
        1 - \delta  & ,i = 1 \\
        1 + \delta & ,i = 2 \\
    \end{cases}
\end{equation}
% \begin{equation}
% \textbf{\emph{T}}_{i,i}(\delta) = 0.6 + 0.3 \cdot Sigmoid(\theta_{i}) + 0.1 \cdot
%     \begin{cases}
%         \delta  & ,i = 1 \\
%         -\delta & ,i = 2 \\
%     \end{cases}
% \end{equation}
where the $S(\cdot)$ is the $Sigmoid$ function, $\theta_{i}$ is the parameter used only for column, and $\theta_{0}$ is the parameter shared by both columns. Adjusting the $\delta$ is equivalent to shifting the geometric space of the simplex, where the spaces with a closer adjustment coefficient share a more common area. This ensures that the distribution of the noise class-posterior has a strong consistency with the adjustment coefficient. Furthermore, the range distribution of any $\textbf{\emph{T}}(\delta)$ is one subspace of an extended transition matrix $\textbf{\emph{T}}^{\Sigma}$, whose edges is defined as the edges of $\textbf{\emph{T}}(-1)$ and $\textbf{\emph{T}}(1)$:
\begin{equation}
\textbf{\emph{T}}^{\Sigma}_{i,i} = 1 + \frac{S(\theta_{0}) \cdot S(\theta_{i}) - S(\theta_{0})}{2}
\end{equation}
% \begin{equation}
% \textbf{\emph{T}}^{\Sigma}_{i,i} = 0.7 + 0.3 \cdot Sigmoid(\theta_{i})
% \end{equation}

To train the proposed ARTran, we organize a group of loss functions to jointly optimize the proposed modules. The benchmark noise posteriors and the adjusted noise posteriors are optimized with classification loss with benchmark and adjusted labels respectively. Following the instance-independent scheme, we optimize the SST by minimizing the volume of the extended SST \cite{li2021provably}. The total loss function $\mathcal{L}$ we give is as follows:
\begin{equation}
\mathcal{L} = \mathcal{L}_{cls}(P(\tilde{\textbf{\emph{Y}}}|x,0),Y(s,0)) + \mathcal{L}_{cls}(P(\tilde{\textbf{\emph{Y}}}|x,\delta),Y(s,\delta)) + \mathcal{L}_{vol}(\textbf{\emph{T}}^{\Sigma})
\end{equation}
where $\mathcal{L}_{cls}(\cdot)$ indicates the cross entropy function, and $\mathcal{L}_{vol}(\cdot)$ indicates the volume minimization function.

\section{Experiments}

\subsection{Dataset}

We conduct experiments on an OCT dataset including 509 volumes of 290 subjects from the same OCT system (RTVue-XR, Optovue, CA) with high diversity in SE range and retinal shape. Each OCT volume has a size of $400$ (frames) $\times$ $400$ (width) $\times$ $640$ (height) corresponding to a $6mm \times 6mm \times 2mm$ volume centered at the retinal macular region. The exclusion criteria were as follows: eyes with the opacity of refractive media that interfered with the retinal image quality, and eyes that have undergone myopia correction surgery. Our dataset contains 234 low (or non) myopia volumes, and 275 high myopia volumes, where labels are determined according to a threshold spherical equivalent $-6.0D$. We divide the dataset evenly into 5 folds for cross-validation according to the principle of subject independence for all experiments. For data selection, we select the center 100 frames of each volume for training and testing, so a total of 50,900 images were added to the experiment. And the final decision outcome of one model for each volume is determined by the classification results of the majority of frames. For data augmentation, due to the special appearance and location characteristics of high myopia in OCT, we only adopt random horizontal flipping and random vertical translation with a range of $[0, 0.1]$.

\begin{table}
\centering
\caption{Comparison of classification methods with benchmark inclusion criteria.}\label{Comparison}
\begin{tabular}{ccccc}
\hline\hline
% \noalign{\smallskip}
    Model &  Parameters($M$) & Accuracy(\%) & Precision(\%) & Recall(\%) \\
% \noalign{\smallskip}
\hline
% \noalign{\smallskip}
    ResNet-50 \cite{he2016deep}                 & 23.5 & 82.97$\pm9.0$ & 85.4$\pm27.4$ & 83.3$\pm4.0$  \\
    EfficientNet-B5 \cite{tan2019efficientnet}  & 28.3 & 81.1$\pm7.0$  & 83.1$\pm14.2$ & 80.4$\pm5.6$  \\
    ViT-Small \cite{dosovitskiy2020image}       & 21.7 & 82.7$\pm4.1$  & 85.0$\pm4.6$ & 82.5$\pm4.7$  \\
    Swin-Tiny \cite{liu2021swin}                & 27.5 & 83.3$\pm13.3$  & 83.9$\pm19.4$ & 85.5$\pm26.1$  \\
    Swin-V2-Tiny \cite{liu2022swin}             & 27.5 & 83.3$\pm7.4$  & 83.2$\pm23.2$ & 86.2$\pm34.7$  \\
    Choi's Method \cite{choi2021deep}           & 47.0 & 83.5$\pm5.7$  & 84.3$\pm12.3$ & 86.2$\pm10.4$   \\
    Li's Method \cite{li2022development}        & 54.3 & 84.8$\pm10.7$  & 85.6$\pm14.5$ & 86.5$\pm30.2$   \\
% \noalign{\smallskip}
\hline
% \noalign{\smallskip}
    ARTran(Ours)         & 21.9 & \textbf{86.2$\pm1.4$}  & \textbf{87.5$\pm2.4$} & \textbf{86.9$\pm2.1$}\\
% \noalign{\smallskip}
\hline\hline
\end{tabular}
\end{table}

\begin{table}
  \centering
  \caption{The ablation study of our anisotropic patch embedding (APE), adjustable class embedding (ACE), and shifted subspace transition matrix (SST).}\label{Ablation}
  \begin{tabular}{ccc|ccc}
   \hline\hline
   % \noalign{\smallskip}
   APE & ACE & SST &  Accuracy(\%) & Precision(\%) & Recall(\%) \\
   % \noalign{\smallskip}
   \hline
    % \noalign{\smallskip}
    \ding{51} & \ding{51} & \ding{51} & 86.2 & 87.5 & 86.9 \\
    \ding{55} & \ding{51} & \ding{51} & 84.1 & 86.6 & 84.7 \\
    \ding{51} & \ding{51} & \ding{55} & 84.7 & 86.5 & 86.2 \\
    \ding{55} & \ding{51} & \ding{55} & 83.9 & 84.8 & 85.6 \\
    \ding{51} & \ding{55} & \ding{55} & 83.3 & 86.2 & 84.4 \\
    \ding{55} & \ding{55} & \ding{55} & 82.7 & 85.0 & 82.5 \\
    % \noalign{\smallskip}
  \hline\hline
  \end{tabular}
\end{table}

\subsection{Comparison Experiments and Ablations}

To evaluate the proposed ARTran in predicting under a benchmark inclusion criteria ($-6.0D$), we compare it with two convolution-based baselines: ResNet-50 \cite{he2016deep} and EfficientNet-B5 \cite{tan2019efficientnet}; three transformer-based baselines: ViT-Small \cite{dosovitskiy2020image}, Swin-Tiny \cite{liu2021swin} and Swin-V2-Tiny \cite{liu2022swin}; and two state-of-the-art high myopia screening methods: Choi's method and Li's method \cite{li2022development}.

Table~\ref{Comparison} presents the classification accuracy, precision, and recall by our ARTran and comparison methods. The proposed ARTran already outperforms baseline methods remarkably with a $2.9\%$ to $5.1\%$ higher accuracy and a lower variance. This is because we design modules to capture the features of high myopia, which brings effectiveness and robustness. Although consisting of fewer parameters, our ARTran obtains higher accuracy than two state-of-the-art screening methods. Moreover, clinical practice requirements generally advocate the model that predicts smaller false negative samples, i.e., a higher recall. It is observed that the recall of our approach is the best performance, which means that our model has minimal risk of missing highly myopic samples. 

We further perform ablations in order to better understand the effectiveness of the proposed modules. Table~\ref{Ablation} presents quantitative comparisons between different combinations of our APE, ACE, and SST. As can be seen, ablation of APE leads to a rapid decrease in recall, which means that this patch embedding approach captures better features about the propensity for high myopia. The ablation of ACE represents the ablation of the proposed adjustment scheme, which leads to lower accuracy. The ACE drives the network to learn more discriminative features for images in the adjustment range during the training process. The ablation of SST leads to a rapid decrease in precision. The possible reason is that the label noise may be more from negative samples, leading to increased false positive samples.

\begin{figure}
\includegraphics[width=\textwidth]{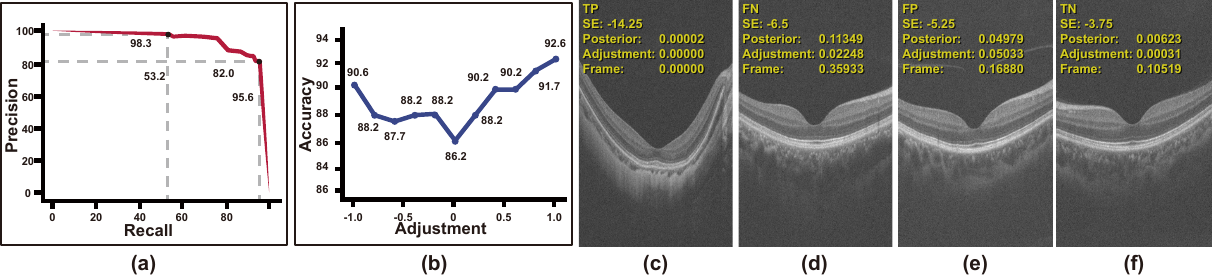}
\caption{Adjustable and uncertainty analyses. (a) PR curve of the adjustment coefficient. (b) Classification accuracy using biased labels with adjustment coefficients. (c) A true positive example. (d) A false negative example. (e) A false positive example. (f) A true negative example.} 
\label{fig-3}
\end{figure}

\subsection{Adjustable Evaluation and Uncertainty Evaluation}

To evaluate the effectiveness of the adjustment module, we change the adjustment coefficient several times during the testing phase to obtain screening results at different thresholds. Fig~\ref{fig-3}(a) depicts the PR curve when adjusting the adjustment coefficient. The performance of the two endpoints ($\Delta=-1$ and $\Delta=1$) is marked. Even with a high recall rate, the precision is not low. Fig~\ref{fig-3}(b) shows the performance of the biased labels for different adjustment coefficients. As can be seen, the accuracy improves when offsetting the inclusion criteria, which on the one hand may be due to the difficulty of classification near the benchmark criteria, and on the other hand proves that the proposed model is effective for the adjustment problem.

To evaluate the interpretability of our ARTran, we propose novel uncertainty scores based on the proposed adjustable scheme: (1) The closer the posteriors closer to 0.5 indicates larger uncertainty. (2) More frames in a volume with disagreement indicate larger uncertainty. (3) Based on a set of adjustment coefficients, the greater the variance, the greater the difficulty or uncertainty. Some examples are shown in Fig~\ref{fig-3}. The two correctly classified (TP\&TN) examples are less difficult and therefore smaller uncertainty score. Large uncertainty scores are more likely to occur around inclusion criteria. Each of the two error examples (FP\&FN) contains at least one uncertainty score higher than the other examples.

\section{Conclusion}

In this work, we proposed ARTran to screen high myopia using OCT images. Experimental results demonstrated that our approach outperforms baseline classification methods and other screening methods. The ablation results also demonstrated that our modules helps the network to capture the features associated with high myopia and to mitigate the noise of labels. We organized the evaluation of the adjustable and interpretable ability. Experimental results showed that our method exhibits robustness under variable inclusion criteria of high myopia. We evaluated uncertainty and found that confusing samples had higher uncertainty scores, which could increase the interpretability of the screening task.

\subsubsection{Acknowledgements} This work was supported by National Science Foundation of China under Grants No. 62172223 and 62072241, the Fundamental Research Funds for the Central Universities No. 30921013105.


\nocite{*}
% \bibliography{ref}

\end{document}